# Automatic Assembly Planning based on Digital Product Descriptions

## Abstract

This paper proposes a new concept in which a digital twin derived from a digital product description will automatically perform assembly planning and orchestrate the production resources in a manufacturing cell. Thus the manufacturing cell has generic services with minimal assumptions about what kind of product will be assembled, while the digital product description is designed collaboratively between the designer at an OEM and automated services at potential manufacturers. This has several advantages. Firstly, the resulting versatile manufacturing facility can handle a broad variety of products with minimal or no reconfiguration effort, so it can cost-effectively offer its services to a large number of OEMs. Secondly, a solution is presented to the problem of performing concurrent product design and assembly planning over the organizational boundary. Thirdly, the product design at the OEM is not constrained to the capabilities of specific manufacturing facilities. The concept is presented in general terms in UML and an implementation is provided in a 3D simulation environment using Automation Markup Language for digital product descriptions. Finally, two case studies are presented and applications in a real industrial context are discussed.

## 1 Introduction

In the last few years, a growing volume of research has emerged on business networks involving manufacturers pursuing efficiency through agility and thus providing a new form of competition to manufacturers with low labor costs (e.g. [6][8][16]). The need for agile manufacturing arises not only from rapidly changing market demand but also from novel design approaches exploring larger parts of the design space [1]. A potential manufacturing concept to address the need for agility is product-centric control, in which the digital counterpart of the product requests its own manufacturing services [3]. However, the research is specific to an industrial partner's business case of a single manufacturer serving several OEMs producing similar but customized products, and the focus is on the logistics level without investigating how the product-centric control can be applied to assembly planning.

Our goal is to address these limitations by proposing a product-centric control concept that enables an OEM to flexibly contact potential manufacturers and involve them already in the product design phase. Such manufacturers will have the ability to support the designers at the OEM to perform virtual manufacturing of their design automatically and promptly according to a digital product description provided by the designer. Our proposed means to achieve this goal is through a new concept of *agility* illustrated in Figure 1, in which digital twin objects generated from the digital product descriptions drive the product-centric control. The digital twin will have the intelligence to perform assembly planning and to orchestrate the production resources in the manufacturing cell. Thus, these resources only need to provide generic services, resulting in great versatility to handle diverse digital product descriptions without a need to reconfigure the cell.

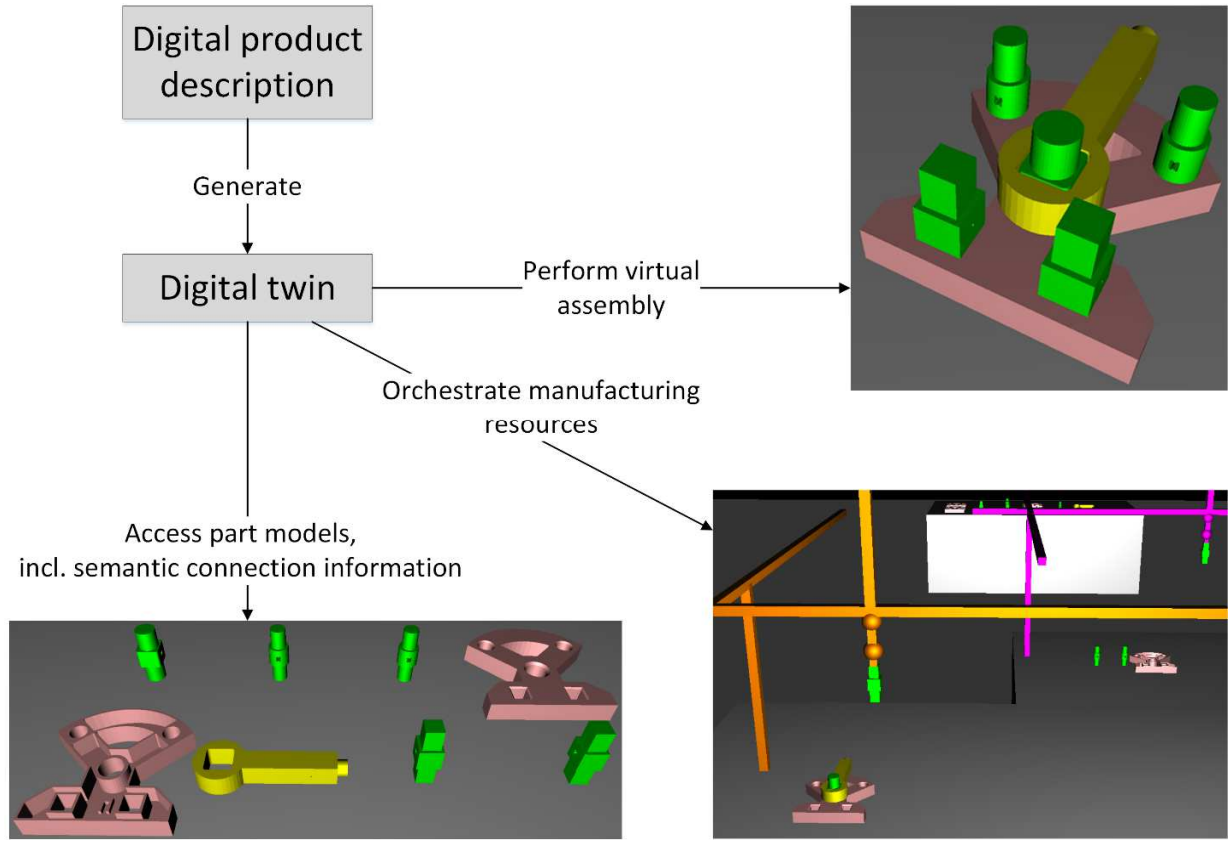

*Figure 1 Concept overview*

The following contributions are made. Firstly, a general framework is presented to realize the concept in shown Figure 1, independent of a particular product description technology. Secondly, an application to the AutomationML product description technology [38][42] is provided. Thirdly, an implementation in a 3D simulation environment is described and demonstrated with two case studies.

## 2 Previous work

Previously the trend in manufacturing industries was to outsource manufacturing operations to low labor cost countries [7]. The more recent trend of Industry 4.0 targets the goal of making manufacturing sustainable in higher labor cost countries. Toward this goal, a number of recent publications on CPS (Cyber-Physical Systems), CPPS (Cyber-Physical Production Systems), Smart Manufacturing and Industrial Internet show a trend towards business networks of companies into which potential participants can offer their specialized services ([6][8][16]). One unresolved challenge for such networks is how they could be performing assembly planning already during the product design phase as this is significant for the efficiency of the eventual manufacturing [18][2], and existing approaches require considerable manual effort, such as using real operators in a virtual assembly environment [19]. To realize the potential of agile business networks, this article investigates how a designer could submit a digital product description to several potential manufacturers with the capability to perform fully automatic virtual assembly planning. Such a capability has not yet been proposed despite numerous manufacturing applications of CPS, Internet of Things (IoT), multi-agent systems, holonic systems and flexible manufacturing systems (e.g. [12][14][15][47]). Our proposal differs from approaches such as product configuration systems [10] and product family architectures [17], whose agility is limited by detailed *a priori* specifications about the range of product to be manufactured.

One specific aspect of CPPS and Industry 4.0 is that data from virtual models of products and production resources is available for optimizing manufacturing operations [9]. In the Industry 4.0 age, approaches for exploiting this data to plan manufacturing operations can be grouped into the more conservative approach of extending MES systems [4][5][47] and to the radical approach of product-centric control. In the latter case, a digital counterpart of the product keeps track of its status and is able to participate in the planning of operations from supply chain management and logistics to routing on the factory floor [26][27][28][29][30][31] exploiting the flexibility from additive manufacturing (AM) [3]. In addition to product-centric control, a similar concept under the name of digital twin has shown active research in 2016-2017 in a number of diverse communities such as CPS, IoT, Industry 4.0 and model-based systems engineering [25][32-36]. Both concepts are similar in the sense that a digital counterpart of the product maintains up to date information and can participate in operations involving some part of the lifecycle of the product. However, the research on digital twins opens new opportunities in the product design phase, in bridging design and operation and especially in 3D simulation of the assembly process [37]. Our proposal is thus extending the product-centric control to the lower level task of assembly planning as well as exploiting the digital twin concept in 3D simulation. Bridging the gap between these bodies of knowledge may lead to the various researches cited above finding broader application in more holistic concepts of agility in manufacturing, spanning supply chain management and product design. Our goal is to define the concept rather than to focus on protocols, service discovery mechanisms and ontologies for which a number of specific works exist [11][13][47][48].

A discussion on agile manufacturing would not be complete without addressing the potential of AM to shorten time to market [24]. As discussed in section 5, after virtual assembly has been performed satisfactorily, a physical assembly validation could be performed using AM to obtain the parts from CAD models provided by the designer. Thus the proposed concept would be poised to integrate with anticipated disruptive developments from AM including decentralized supply chain management [20] and hybrid approaches combining AM and conventional technology for reduced time to market and investment risk [21]. Although current costs for AM limit its application [22], decreasing costs and increasing market demand for customization result in conventional manufacturers facing increasing competition from entrants with AM technology [23].

Assembly planning is the process of creating a sequence of assembly motions to craft a product from separate parts considering the final product geometry and the production environment including resources and obstacles. Assembly planning can be divided into Assembly Sequence Planning (ASP), which considers the parts as free flying, and Assembly Path Planning (APP) that considers the trajectories those parts move along and the relevant manipulators. Automatic methods for ASP have been studied since 1980's [56, 57, 58] and the study of robotic manipulation planning continues actively in the framework of integrated task and motion planning as the manipulation planning problem contains both discrete (which part next) and continuous (which continuous trajectory) components [59,60]. The planning is often hierarchical such that the geometric decisions are postponed (e.g. [61]). The continuous geometric planners involve often stochastic search approaches (e.g. [62]) using a simulation model for collision checking. The concept proposed in this paper is compatible with the current literature such that current ASP and APP methods can be used to implement those functionalities within the framework.

# 3 Conceptual framework

This section proposes a conceptual framework for the key entities of a manufacturing cell performing product-centric control down to assembly planning based on digital product descriptions. The framework is presented in UML, since as has been discussed in section 2, numerous alternative technologies, studies, standards, ontologies and interfaces exist related to various aspects of the framework. UML and its SysML extension have been applied to a hierarchical decomposition of industrial application data models [43] and to the mechatronic product design process [46], so assembly planning is a next step. This section does not aim to provide an exhaustive discussion of all aspects of the UML diagrams. A thorough explanation is provided in the supplementary materials document "UML_detailed_description".

Figure 2 shows the key classes and some of their members that are most relevant for this discussion. At first glance, it is evident that the DigitalTwin, which is instantiated from the digital product description, occupies the most central position. The purpose of this design is to have all product specific data and functionality in the DigitalTwin and in the classes on the left side of the figure, so that the Cell and all the classes under it only provide simple generic functionality, resulting in the kind of agile cell proposed in section 1. Through its association to the Cell class, the DigitalTwin is able to gain access to all production resources and to orchestrate them, thus realizing our proposed product-centric control concept. It is notable that the Connection class provides semantic information about which parts are actually connected as opposed to being adjacent – this information is not available in plain CAD models. Figure 2 has significant detail, but abstracts away all information related to the product description technology and the 3D simulation or real manufacturing cell automation systems.

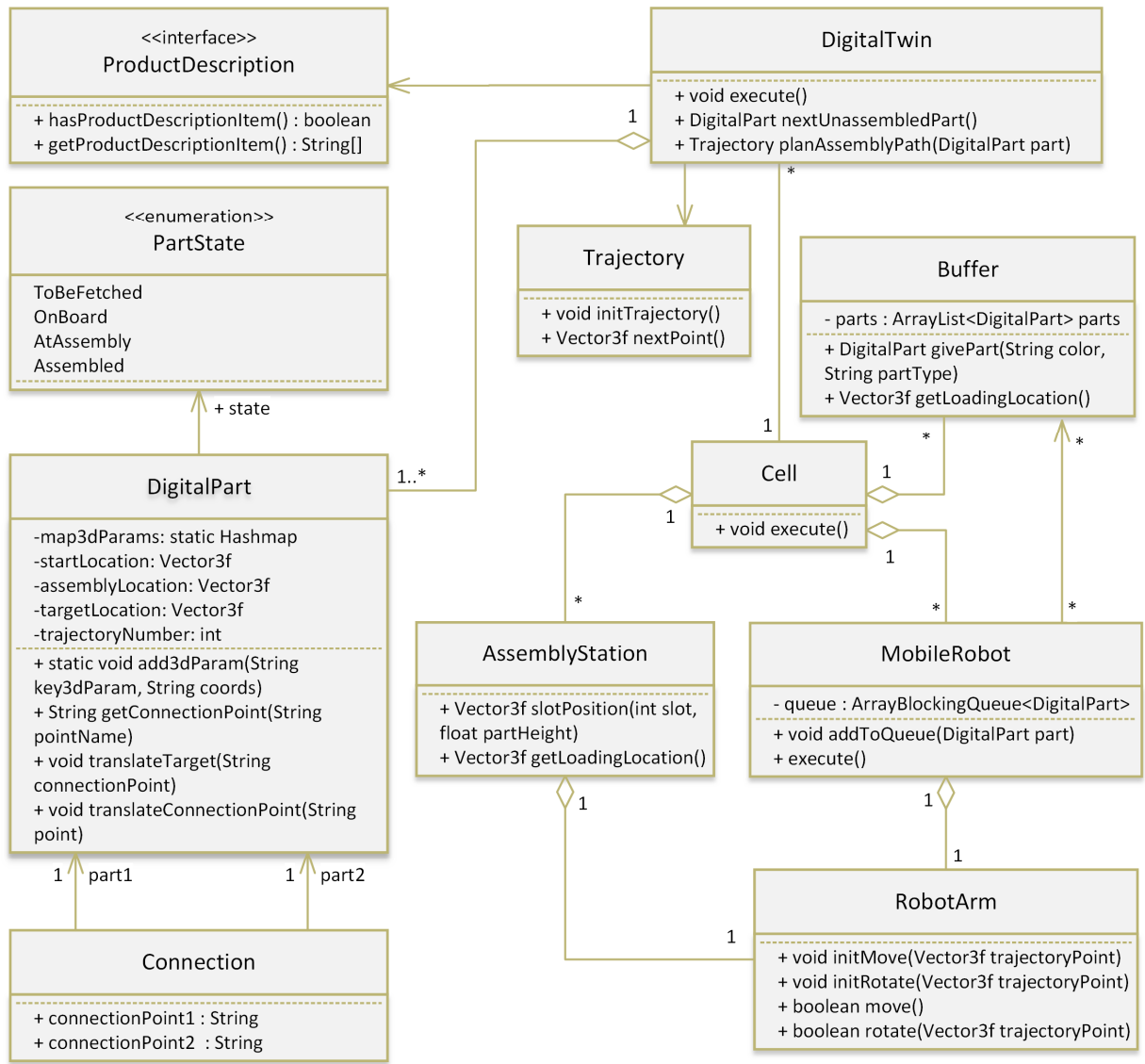

*Figure 2 Key classes in the framework*

Figure 3 shows the sequence diagram for reading the digital product description. UML2 combined fragments are used for the loop and alternatives. As can be seen from Figure 2, ProductDescription is an interface for which a separate implementation should be made for each supported digital product description format. The ProductDescription class processes the digital product description and forms a list of items, each of which is encoded as an array of strings. The first element in this array should be the "create", "parameter" or "connection" used in Figure 3 to create a DigitalPart, a parameter or a Connection between two DigitalPart objects. The parameters are 3D coordinates of DigitalPart's mechanical interfaces, so the parameter applies to all DigitalPart objects of a specific component type. As can be seen from Figure 2, the add3dParam() method adds the parameter to a static HashMap object. The keys of the HashMap are formed by appending the parameter name to the component type name, thus avoiding any ambiguity issues that might otherwise arise from using a static HashMap.

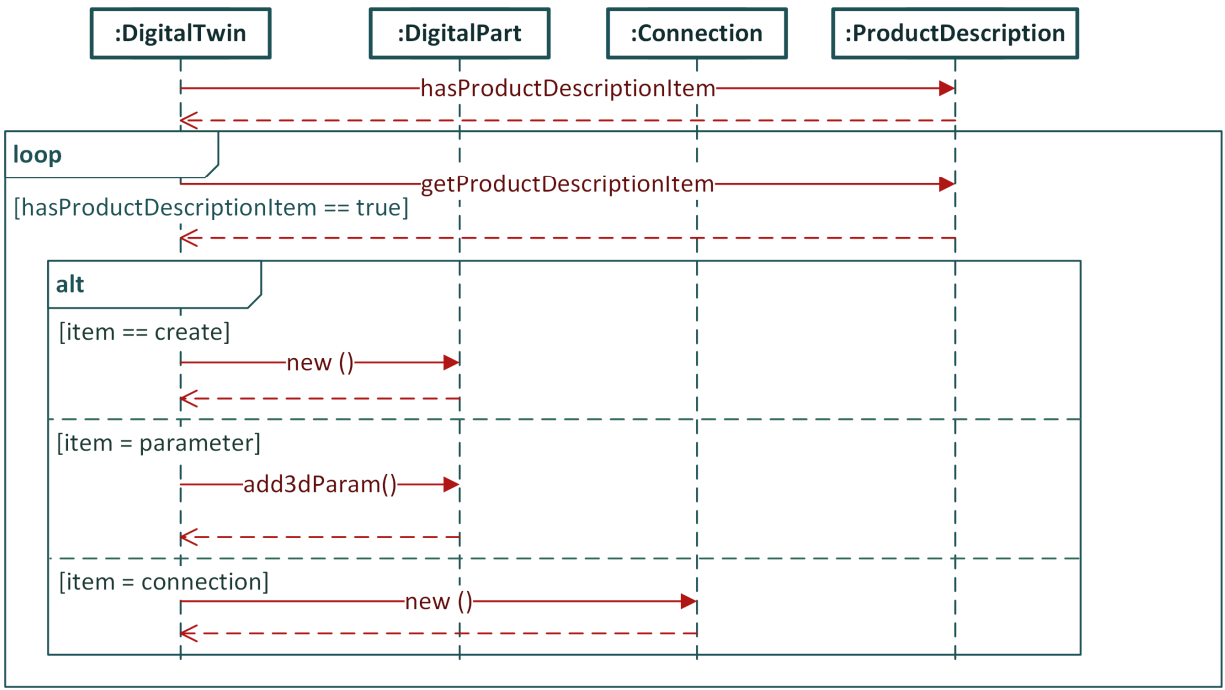

*Figure 3 Reading the digital product description*

Figure 4 shows the process of creating the assembly, resulting in the 3D coordinates of each part in the final assembly. Before this algorithm is executed, one part is selected arbitrarily, translated to the origin and then its "assembled" attribute is set true. The algorithm in Figure 4 has an outer loop that iterates through the collection of Connection objects until it finds one in which "part1" is assembled and "part2" is not assembled (please refer to the "part1" and "part2" associations of the Connection class in Figure 2). How this connection is made is described in the "opt" combined fragment. From the Connection object "c", the string name of the connection point of "part1" (which was obtained from the digital product description) is passed to the getConnectionPoint() method of part1, which returns a 3D coordinate point using the static HashMap of the DigitalPart class. The hitherto unconnected "part2" is now translated, using the "translateTarget" method, to this connection point of the already assembled "part1". Now the center of part2 is at the connection point of part1, so it is necessary to perform one more translation so that the relevant connection point of "part2" (as specified in the digital product description and parsed into the attribute "c.connectionPoint2") is translated by "translateConnectionPoint" to the said connection point of "part1".

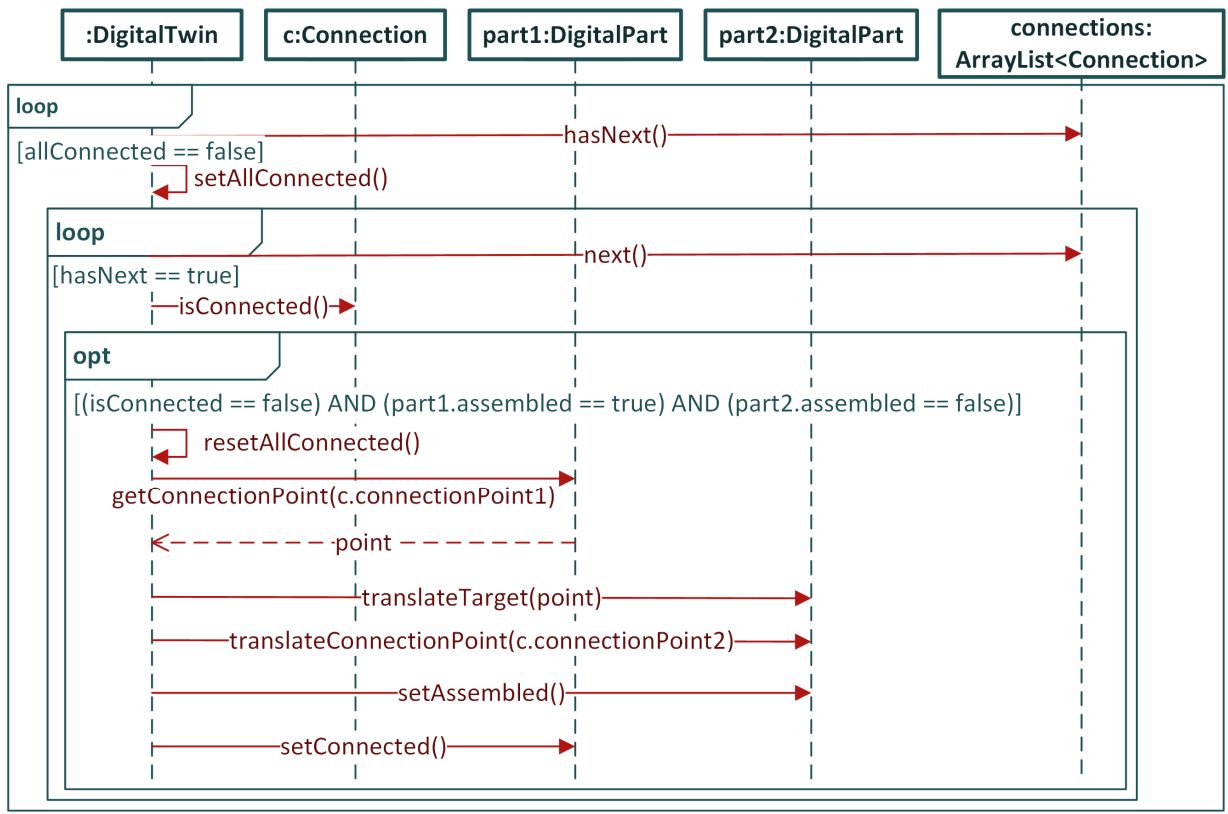

*Figure 4 Creating the assembly*

Now the 3D coordinates of all parts in the final assembly are known and can be used for ASP by DigitalTwin's "nextUnassembledPart" method, which has the following proof-of-concept implementation: a list is made of the unassembled parts that will be placed on the assembly surface, and if the list is not empty, one arbitrarily selected part will be returned. Otherwise, a list is made of all unassembled parts that have a connection to an assembled part. From this list, the part with the lowest height coordinate is returned. As has been discussed in section 2, more sophisticated ASP algorithms have been published, and the goal of this paper is not to prose new such algorithms but rather a design where such algorithms can be inserted as alternative implementations of the "nextUnassembledPart" method. During virtual assembly, the returned DigitalPart object is added to the queue of the MobileRobot (see Figure 2), so that during physical assembly, parts will be delivered to the assembly station in the order in which they will be assembled.

3D simulation environments typically have a cyclically executed update method that renders the screen graph, i.e. draws onto the 2D screen all 3D objects that are visible through the camera. The sequence diagram in Figure 5, responsible for virtual assembly planning and physical assembly, is called cyclically from this update method. Thus, over the course of assembling one DigitalPart object, the sequence is executed numerous times and the guards for the opt, alt and break fragments control which messages are sent in one particular execution; a detailed explanation is provided in the supplementary materials document "UML_detailed_description". Figure 5 uses exclusively synchronous messaging, implying a single-threaded implementation, which can be used in an automation architecture with central control.

An expert in distributed automation will be able to modify the sequence with asynchronous messaging if a decentralized event-based control is desired. The multi-agent control paradigm for decentralized control architecture can be applied following, for example [50-52]. It was exemplified in [53-55] how the control logic could be organized in a purely decentralized way and executed distributed on the Industrial Internet of Things based hardware infrastructure in the automation context using the IEC 61499 architecture.

The functionality of Figure 5 is available in existing assembly planning tools, and the contribution has been to incorporate this functionality into the product-centric control framework of Figure 2. Everything until the call to "nextUnassembledPart" is related to the simulation of assembling the "part" object along a trajectory "t". "nextUnassembledPart" performs one step of ASP and returns the "part" object in Figure 5. Subsequently, APP for this part is performed in the "planAssemblyPath" method. "planAssemblyPath" has a proof-of-concept implementation for downwards assembly and some special cases where the part needs to be connected upwards. In the general case, APP returns a Trajectory object ("t" in Figure 5) which has a sequence of translational and rotational movements. More sophisticated APP algorithms have been discussed in section 2 and may be added as alternative implementations of the "planAssemblyPath" method. If a collision is detected, another call to "planAssemblyPath" results in a new trajectory or a fallback to ASP, in which "nextUnassembledPart" may perform a partial disassembly and then proceed with another assembly sequence. The fallback is not included in our proof-of-concept implementation.

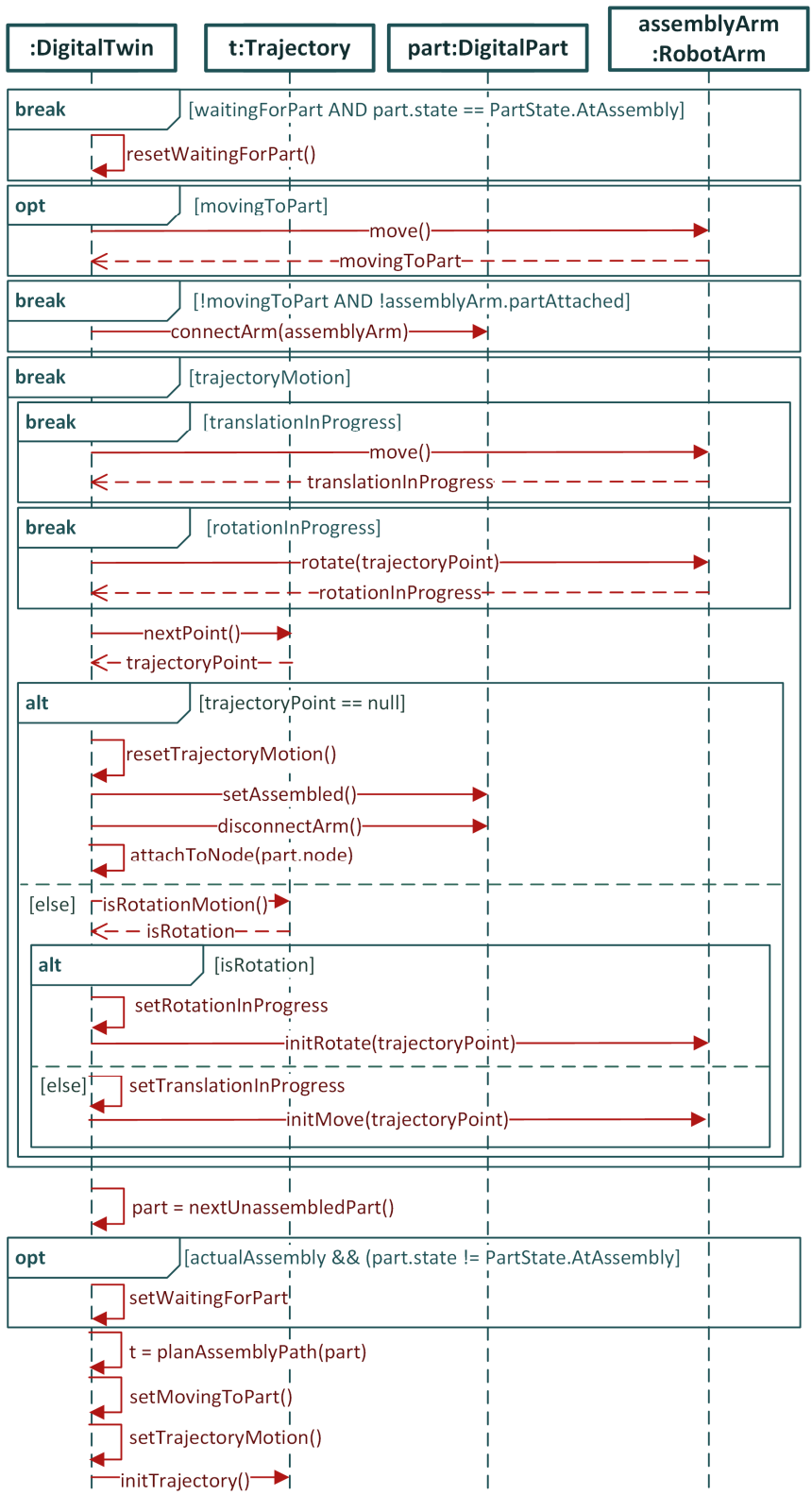

*Figure 5 Virtual assembly (if "actualAssembly" is false) or physical assembly (if "actualAssembly" is true)*

As can be seen from Figure 5, the product-centric control has the DigitalTwin object orchestrating the assembly robot, sending it commands to translate, rotate, connect and disconnect the part. In physical assembly, the first break fragment pauses the assembly robot if the part has not yet been delivered to the assembly station by the mobile robot. The mobile robot has also been under the control of the DigitalTwin, which has filled the queue of parts to be transported during the ASP phase. The mobile robot in turn queries the buffers for those parts. The buffer's "givePart" method (see Figure 2) provides a DigitalPart object if a suitable part is present in the buffer, and DigitalPart has the 3D location information for the MobileRobot's pick operation. The route planning of the MobileRobot is simple: all buffers and the assembly station are visited in sequence, but more sophisticated algorithms are available [44][45]. Thus the assembly station, the transportation system and the buffers have no hardcoded assumptions about what types of parts are processed or what kinds of ASP or APP should be performed. This intelligence is built into the DigitalPart, Trajectory, Connection and DigitalTwin objects automatically from the digital product description according to the algorithms in Figure 3, Figure 4 and Figure 5, thus realizing the concept of agility proposed in section 1.

## 4 Implementation

Several digital product description technologies are available. All CAD and PDM/PLM (Product Data/Lifecycle Management) systems have an internal format, and standards to exchange the information exist. For our purposes, an ideal technology is lightweight, open, standardized and XML based, so that it can be processed with standard W3C technologies. One such technology is AutomationML (Automation Markup Language) and especially the part of it that uses CAEX (Computer Aided Engineering Exchange) to describe the logical structure of products or production facilities. CAEX and AutomationML were introduced to the international academic community in 2008 [38][42]. Recent activity relevant to our proposed concept includes support from major tool vendors [40], securing the transmissions of the AutomationML content [41] (especially important for our idea of exchanging digital product descriptions from OEMs to potential manufacturers) and ongoing development of the CAEX standard [39]. The method in Figure 3 has been implemented with AutomationML and described in this section for proof-of-concept purposes. AutomationML gives significant freedom for the user to add all of the required information such as the 3D coordinates of the mechanical interfaces of the parts, so this section provides enough detail for an interested reader to replicate the setup.

In order to apply the concept in section 3 to a specific digital product description technology, the following information from Figure 2 needs to be captured: the data for DigitalPart objects, Connection objects and 3D parameters for the DigitalPart types. Further, an implementation needs to be provided for the ProductDescription interface in Figure 2 and Figure 3. An XSLT-transformation (Extensible Stylesheet Language Transformation) was developed that parses the CAEX document and extracts the items in Figure 3 to a text file, which was then read by a Java class implementing the ProductDescription interface in Figure 2 and Figure 3. The XSLT file is available in supplementary material: AutomationML_XSLT.

In this paper, digital product descriptions have been created with the free AutomationML Editor tool available from www.automationml.org , which saves the digital product description in an .aml file which is a regular XML file conforming to the CAEX schema that can be viewed with any XML editor or text editor. The Cranfield benchmark case study, which is elaborated in section 5.2, will be used to illustrate the application of AutomationML with snippets taken from the AutomationML file "Cranfield" which is provided as supplementary material. Figure 6 shows several parts that are referred to in the following descriptions, including the pink faceplate back, yellow pendulum, green shaft in the center, angular bolts in the foreground and round bolts in the background.

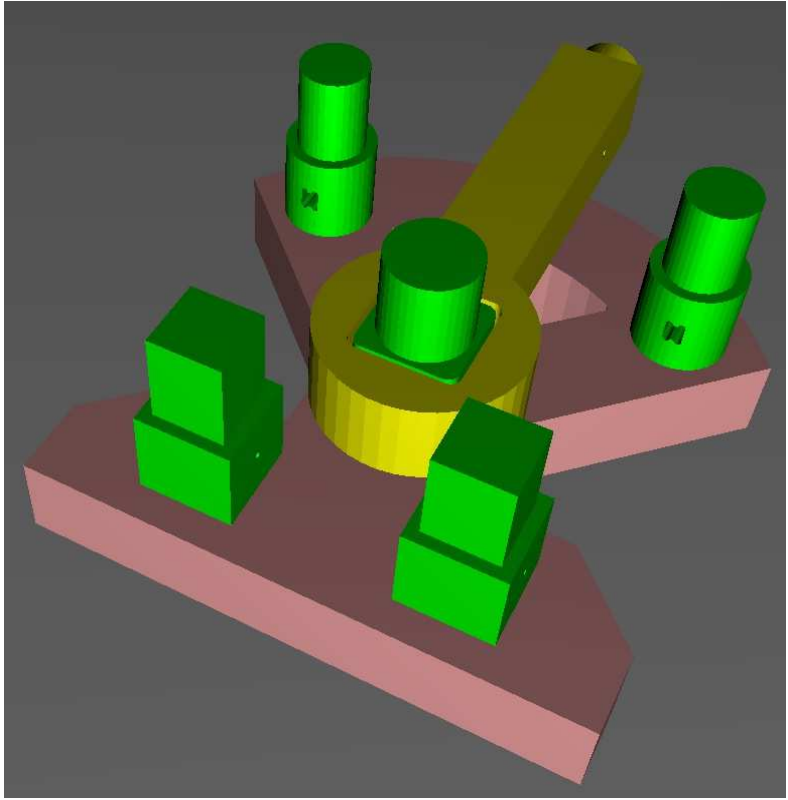

*Figure 6 The Cranfield benchmark case study before the last part (the faceplate) is assembled*

Full details of the CAEX schema are available in IEC 62424 [49], but this presentation is aimed at giving the reader sufficient understanding of the schema without having to consult the standard. All italic names in this section refer to elements defined in the CAEX schema. CAEX consists of 4 parts: *InterfaceClassLib*, *RoleClassLib*, *SystemUnitClassLib* and *InstanceHierarchy*. *InterfaceClassLib* can be used to specify any kind of interfaces, including interfaces between CAD designed parts, such as holes in the faceplate where the angular bolts in the foreground of Figure 6 may be inserted (IBoltAngular in the following snippet). The following XML snippet has the *InterfaceClassLib* for our Cranfield benchmark case study:

```
<InterfaceClassLib Name="InterfaceClassLib1">
        <Version>1.0.0</Version>
        <InterfaceClass Name="IBolt" />
        <InterfaceClass Name="IBoltAngular" />
        <InterfaceClass Name="IShaft"></InterfaceClass>
        <InterfaceClass Name="IPendulum" />
</InterfaceClassLib>
```

*RoleClassLib* and *SystemUnitClassLib* are used to describe components types, so the vendor-independent components are placed in the former and vendor specific components in the latter. Our case studies do not involve a multi-vendor scenario, so all component types are defined in the *SystemUnitClassLib* using the interface definitions from *InterfaceClassLib* and adding *Attribute* elements such as color and orientation of the part in the assembly. These *Attribute* elements will have a unique value for each instance of this *SystemUnitClass*. There are also other kinds of *Attribute* elements for which the value will be same for all instances, namely the 3D coordinates of the mechanical interfaces, so these coordinates are specified in the *DefaultValue* in a string format that the implementation of ProductDescription can parse into the Vector3f 3D coordinate data type used in Figure 2. From each such *Attribute*, a "parameter" item is generated in Figure 3 containing the Vector3f type coordinate and a HashMap key consisting of the *SystemUnitClass* name and the *Attribute* name; this key and coordinate can be given as arguments to the "add3dParam" method of DigitalPart in Figure 2. The following XML snippet has the definition of one *SystemUnitClass*, FaceplateBack, which has the above-mentioned color and orientation *Attribute* elements as well as 5 other *Attribute* elements with 3D coordinates in their *DefaultValue*. There are also 5 *ExternalInterface* elements typed by *InterfaceClass* elements from the *InterfaceClassLib*. Their *Name* matches with the *Name* of the *Attribute* that has the 3D coordinates of the interface in its *DefaultValue*.

```
<SystemUnitClass Name="FaceplateBack">
        <Attribute Name="color" AttributeDataType="xs:string">
                <Description>color</Description>
        </Attribute>
        <Attribute Name="orientation" AttributeDataType="xs:string">
                <Description>orientation</Description>
        </Attribute>
        <Attribute Name="square_left">
                <DefaultValue>-0.656,0,1.151</DefaultValue>
        </Attribute>
        <Attribute Name="square_right" AttributeDataType="xs:string">
                <DefaultValue>0.664,0,1.151</DefaultValue>
        </Attribute>
        <Attribute Name="circle_left" AttributeDataType="xs:string">
                <DefaultValue>-1.021,0,-0.774</DefaultValue>
        </Attribute>
        <Attribute Name="circle_right" AttributeDataType="xs:string">
                <DefaultValue>1.019,0,-0.774</DefaultValue>
        </Attribute>
        <Attribute Name="shaft" AttributeDataType="xs:string">
                <DefaultValue>-0.006,0,0.226</DefaultValue>
        </Attribute>
        <ExternalInterface Name="circle_left"
RefBaseClassPath="InterfaceClassLib1/IBolt" ID="5c244537-0fea-46a5-bf81-7c81252e232e"
/>
        <ExternalInterface Name="circle_right"
RefBaseClassPath="InterfaceClassLib1/IBolt" ID="041ad697-85aa-4c55-8f74-f81c279655b2"
/>
        <ExternalInterface Name="shaft" RefBaseClassPath="InterfaceClassLib1/IShaft"
ID="b430fdb5-6397-438a-b080-1d0f41d67b9e" />
        <ExternalInterface Name="square_left"
RefBaseClassPath="InterfaceClassLib1/IBoltAngular" ID="91946ab4-5e02-4c2c-b90a-
6484c342bfcc" />
```

```
        <ExternalInterface Name="square_right"
RefBaseClassPath="InterfaceClassLib1/IBoltAngular" ID="5c0dd7bb-adee-423f-8d69-
7387a65d05d1" />
</SystemUnitClass>
```

The *InstanceHierarchy* has *InternalElement* elements, each of which corresponds to a DigitalPart object, so the ProductDescription interface's implementation generates a "create" item as in Figure 3 with attributes including the type (i.e. the *SystemUnitClass*), the instance name, as well as the above-mentioned color and orientation attributes. The following XML snippet has the first lines of the *InstanceHierarchy*. There is one *InternalElement* with *Name* "CranfieldBenchmark", and all other *InternalElement* elements are children of it. The snippet includes the first few lines of the *InternalElement* with *Name* "back". The value of *RefBaseSystemUnitPath* refers to the *SystemUnitClass* description in the previous snippet. The only *Attribute* that is shown is "color" which is here assigned the value "pink". *ID* is a RFC4122 UUID (Universally Unique IDentifier) automatically generated by the AutomationML Editor.

```
<InternalElement Name="CranfieldBenchmark" ID="bbe6b4c5-43ad-43cc-9cf8-594fe7e255bb">
        <InternalElement Name="back" RefBaseSystemUnitPath=
"SystemUnitClassLib1/FaceplateBack" ID="2ddc2cf5-1127-471f-b174-dc5b088c9d23">
                <Attribute Name="color" AttributeDataType="xs:string">
                        <Description>color</Description>
                        <Value>pink</Value>
                </Attribute>
```

Finally, the CAEX needs to include information about logical connections between components in order to create the Connection objects in Figure 2 and Figure 3. This information is specified in *InternalLink* elements. The following code snippet is part of the *InternalElement* with *Name* CranfieldBenchmark that was introduced in the previous snippet. It states that the "square_left" interface of the *InternalElement* with *Name* "back" is connected to the "bottom" interface of the *InternalElement* with *Name* "boltA1". The value for *RefPartnerSideA* and *RefPartnerSideB* is a string that is obtained by appending the following: the *ID* of the *InternalElement*, the character ':' and the said name of the interface.

```
<InternalLink Name="InternalLink1" RefPartnerSideA="2ddc2cf5-1127-471f-b174-
dc5b088c9d23:square_left" RefPartnerSideB="2778cfbb-0567-47a4-938a-
bfcb5cdec422:bottom" />
```

The framework described in section 3 has been implemented in the Java 3D simulation environment JMonkeyEngine3, which provides functionality such as translational and rotational movement, importing of CAD and collision detection based on bounding boxes and spheres. When assembling box-like parts such as Lego blocks, this results in satisfactory detection of collisions between the robot arm, a part attached to the arm and the existing assembly. For proof of concept, collision detection is demonstrated in the Lego tower case study. Collision detection of arbitrary CAD objects has not been implemented, but is a standard feature in 3D mechanics simulators such as Bullet. For the same reason, collision detection between the assembly robot arm and the mobile robot arm has not been implemented. The implemented robot is a Cartesian robot with an additional two rotating joints, so that it can rotate the parts and perform upwards assembly of parts with suitable dimensions such as Lego blocks.

# 5 Results and discussion

## 5.1 Lego tower case study

The concept in section 3 and its implementation in section 4 result in a manufacturing cell that can flexibly manufacture a high number of possible products from square and rectangle legos. The range of possible products is less explicitly constrained than in the case of other approaches such as product configuration systems [10] and product family architectures [17]. The AutomationML digital product description of one example Lego product is the "Lego" file provided in supplementary material. Figure 7 shows a screenshot of the assembled product. The grey surface is the AssemblyStation from Figure 2 and the orange mechanism is its RobotArm, which has just performed an upwards assembly of the last pink lego. Figure 8 shows a screenshot from the virtual assembly phase where the robot arm has collided with the assembly when it was attempting an assembly into the direction of the negative X-axis. It will then attempt another assembly into the direction of the positive X-axis and succeed.

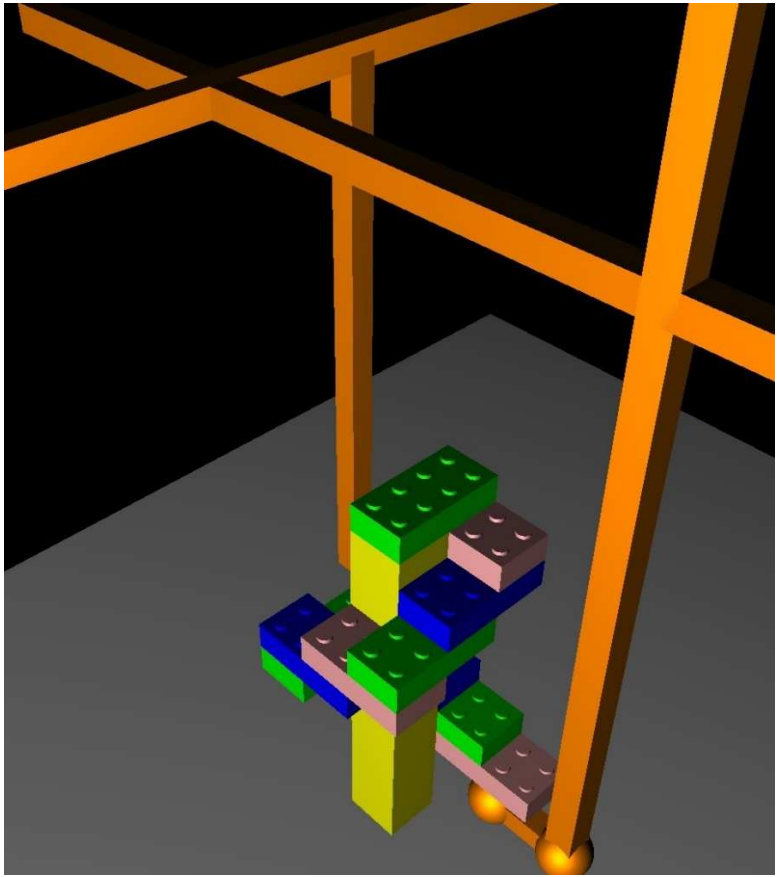

*Figure 7 The assembled lego example and the assembly station's robot arm*

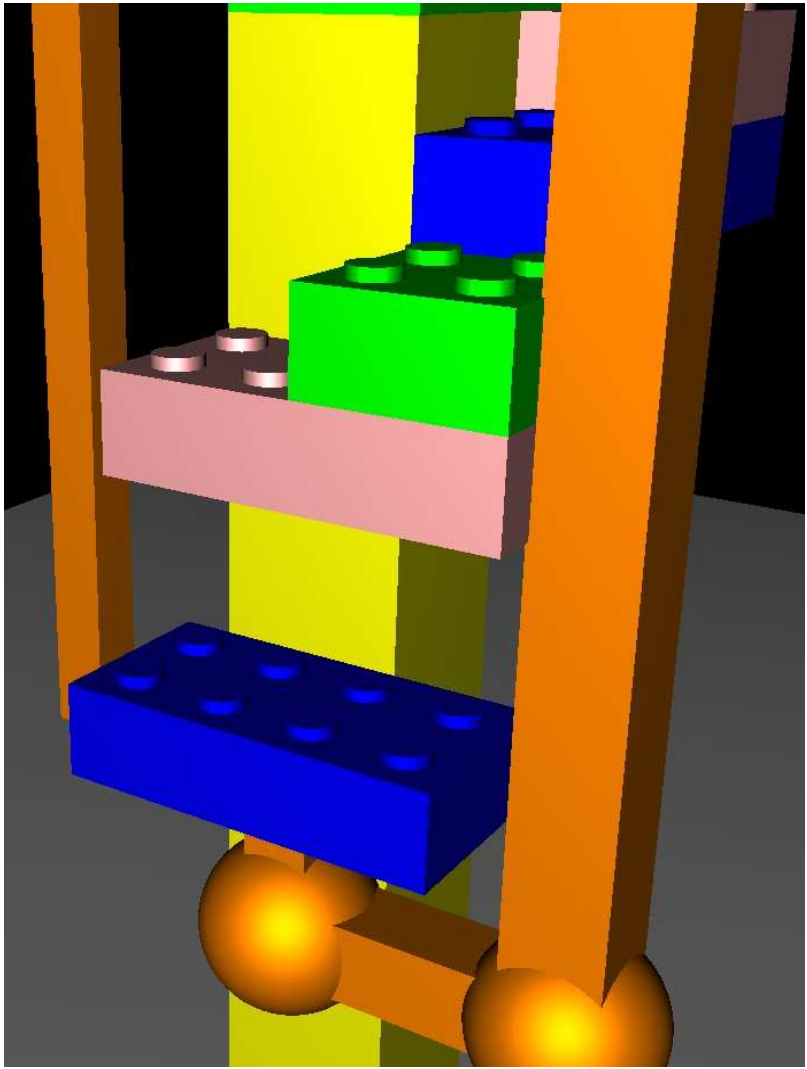

*Figure 8 Collision between the robot arm and assembly*

Figure 9 shows a screenshot of the physical assembly phase, including the following objects introduced in Figure 2 which were not visible in previous screenshots: the MobileRobot platform with a magenta RobotArm and two Buffer objects in the background: one for rectangular block on the left and another for square block on the right.

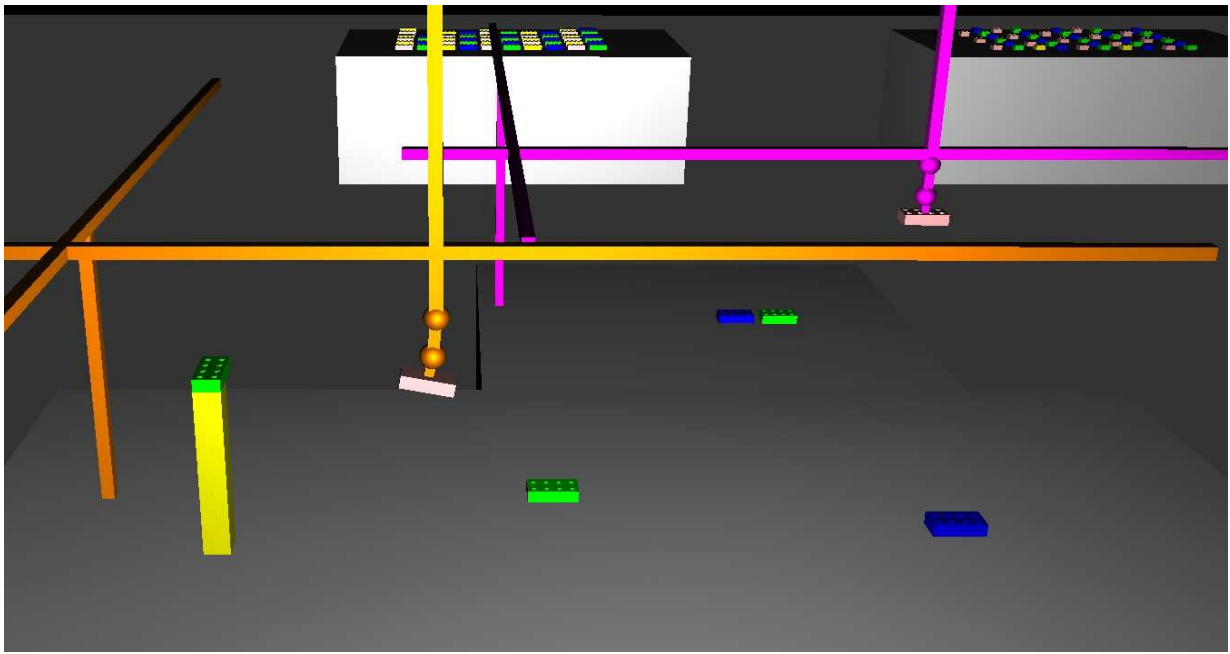

*Figure 9 Screenshot from the physical assembly phase showing the entire manufacturing cell.*

[INSERT LINK TO VIDEO FILES LegoTower_Android.mp4, LegoTower_iPhone.mp4

Caption: Video of virtual and physical assembly for the lego tower case study]

A video file titled LegoTower_Android.mp4 is included as an electronic attachment, which has been tested with Windows computer and Android Phone. Another video with exact same content is provided with name LegoTower_iPhone.mp4, which has been tested with iPhone and MacOS. The video starts after the procedures in Figure 3 and Figure 4 have been completed. The video illustrates the behavior captured in Figure 5, first for virtual assembly and then for physical assembly. As has been justified in section 4, collision detection between the assembly robot arm and the mobile robot arm has not been implemented. The reader is strongly encouraged to view the video, since figures can only provide a very partial and static view into the workings of the implemented concept.

## 5.2 Cranfield benchmark case study

The framework in section 3 is not specific to legos. Any CAD models can be used for the DigitalPart objects, which has an implementation that only needs to be provided the name of the CAD model, which has been preprocessed and imported to the 3D simulation environment. The preprocessing involves scaling, translation and conversion from CAD format to Wavefront (.obj) format, which can be performed in a CAD or 3D content creation tool, including free tools such as FreeCAD and Blender.

The Cranfield assembly benchmark [63] is chosen as our second case study, but the implementation can handle other designs of similar complexity, and the concept in section 3 can handle more difficult assemblies when the current assembly planning implementation is replaced for example with some of the solutions referenced in section 2. Figure 10 shows a sequence of screenshots. As described in section 3, the implemented ASP uses the heuristic of assembling the lowest part with a connection to an already assembled part. Thus in the top-left, the green shaft is assembled to the faceplate, followed by the yellow pendulum in the top-right. In physical assembly, it is important to insert the shaft before inserting the pendulum, since inserting the shaft after placing the pendulum would have much lower margin for error. However, the undesirability of the latter sequence will not be evident in the 3D simulation environment, since no collision will occur. In our implementation, the "Cranfield" file has been designed in such a way that the shaft has a CAEX InternalLink to the faceplate and the pendulum has an InternalLink to the shaft, so the ASP algorithm in section 3 will ensure that the pendulum can only be assembled after the shaft is in place. This is one advantage of using CAEX instead of plain CAD files: the semantics can be used to distinguish connected parts from adjacent parts, with implications to ASP. In the bottom-right of Figure 10 the remaining bolts are assembled only after the pendulum, but this ordering was not constrained by the CAEX but is rather an implementation detail of the chosen heuristic for ASP. In the bottom-right, the top faceplate was assembled only after all five green parts were assembled – this ordering was not fully constrained by the CAEX; rather it emerges as the result of the CAEX definitions and the ASP solution, implying that the product designer should have awareness of ASP.

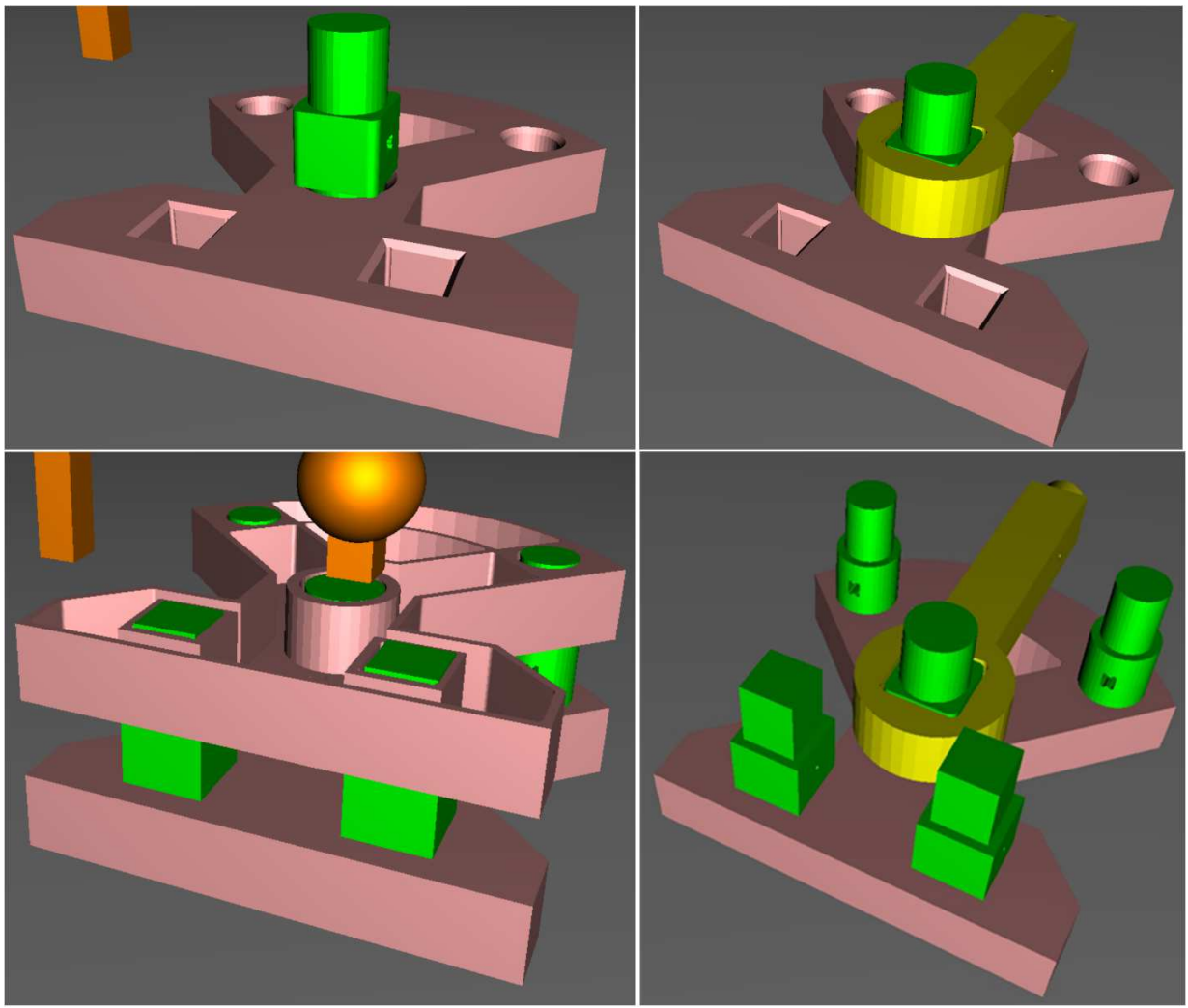

*Figure 10 Sequence of screenshots from the assembly of the Cranfield benchmark: clockwise starting from top-left*

As discussed in section 1-2, the designer is expected to influence the assembly sequence by a well-designed digital product description and then send it to a potential manufacturer for virtual assembly, which can provide video feedback capturing findings such as in Figure 10. Currently, the only manual engineering work from the manufacturer is a few mouse clicks to import the AutomationML and CAD files and to create the video, but software professionals could fully automate this process without further research, so that the designer could get the response in minutes.

A CAD model includes information about what parts are adjacent to each other, but it does not explicitly specify which parts are assembled together and what are the mechanical interfaces by which they should be connected. For example, in Figure 10 the pendulum is adjacent to the faceplate and the shaft, so the internal links in the product description in section 4 are examined to determine that the shaft is assembled first. A well-designed product description has such linkages to ensure that the design is practical from the assembly perspective. It is not expected that the product designer can specify such a description on the first attempt. One application of the proposed system is to provide rapid video feedback to the designer regarding how a specific production facility would assemble the design. The designer can update the linkages in the product description to constrain the ASP.

[INSERT LINK TO VIDEO FILES Cranfield _Android.mp4, Cranfield _iPhone.mp4

Caption: Video of virtual and physical assembly for the Cranfield benchmark case study]

A video file titled Cranfield _Android.mp4 is included as an electronic attachmentand has been tested with Android Phone and Windows. Another video with exact same content is provided with name Cranfield_iPhone.mp4, which has been tested with iPhone and MacOS. The video starts after the procedures in Figure 3 and Figure 4 have been completed. The video illustrates the behavior captured in Figure 5 starting with the virtual assembly and then continuing to the physical assembly.

Since all DigitalPart objects have been created from CAD files, it is possible to use AM technology to physically obtain these parts at the manufacturing facility. The framework in section 3 is equally applicable to a 3D simulation environment and a real environment. Our implementation includes details specific to 3D simulation which would need to be replaced with code that interfaces to the real production resources. The robotics related code would also need to be updated to handle real world robotics issues that can be ignored in a virtualized environment.

## 5.3 Generalization to other cases

The case studies in sections 5.1 and 5.2 are based on a proof-of-concept implementation of the system defined in section 3; the implementation includes a simple ASP and APP solution that is able to manufacture a wide range of products with the constraint that the parts are inserted to the assembly with a vertically downward or upward movement. This is obviously a serious limitation for real products. However, any other ASP or APP implementation may be substituted without changes to the UML specifications in section 3, since the change is limited to the implementation of the methods "nextUnassembledPart" and "planAssemblyPath" of the "DigitalTwin" class, as explained in section 3. A more realistic implementation could be achieved by applying methods from knowledge representation and reasoning for combining ASP and APP, e.g. as in the KnowRob framework [64]. An APP solution e.g. with OMPL (Open Motion Planning Library) [65] can be used to test the feasibility of a particular sequence; as long as the test results are negative, the ASP will continue to generate alternative sequences.

However, there is one limitation towards generality, which would require an extension to the specifications in section 3. Currently the range of ASP solutions is limited by the assumption that parts are added to the assembly one at a time. In order to support ASP involving subassemblies, the sequence in Figure 5 would need to be executed for each subassembly. For products with several subassemblies involving other subassemblies, this raises significant research challenges about the optimal ordering of the subassemblies, which would require modelling aspects related to enterprise resource planning and manufacturing operations management beyond the scope of this paper. Thus this problem is left for further research; the solution to this problem can result in establishing a stronger linkage to the existing body of research on product centric control on supply chain management and logistics, referenced in section 2.

Further limitations to the specifications in section 3 will be encountered for assemblies with require tools e.g. for screwing. However, such an extension would be possible in further work, if the required tool for the assembly operation can be determined based on the type of the part to be assembled: for example, a screw can be assembled by screwing. In order to achieve industrywide standard solutions, catalogue standards such as eCl@ss should be applied; efforts are underway [66] for integrating eCl@ss and the digital product description technology AutomationML used in section 4.

# 6 Conclusion

This research has been motivated by the observation that much of the literature cited in sections 1-4 has paved the way for disruptive developments in agile manufacturing networks, resulting in quick and inexpensive mechanisms for OEMs and potential manufacturers to investigate partnerships in the product design stage. This paper has proposed a concept for this purpose, involving an automated virtual assembly that the manufacturer could perform automatically exploiting semantic part connections information in the product descriptions. Further extension of this concept could lead to a collaborative and interactive network-centric process of product design and manufacturing planning, where manufacturing constraints would lead to product modification as it is being designed.

In order to bring the proposed concept to real industrial applications, further work is required. Although much relevant work for service discovery, integration mechanisms, architectures and interfaces exists, it needs to be adapted to the proposed framework. The implemented robotic solution in the 3D simulation environment also needs to be upgraded using libraries for ASP and APP as required by the type of product in question.

The proposed concept also opens new opportunities for product-centric control in the area of supply chain management and logistics. It is now possible for agents at the OEMs and manufacturers to determine which are the potential manufacturers and to then identify the manufacturer with which the OEM can obtain the most profitable partnership according to criteria such as lowest time to market and cost. In this case, the partnering would be driven by the designer at the OEM performing concurrent product design and assembly planning over the organizational boundary with several potential manufacturers.

## Acknowledgement

The research was supported in part by Academy of Finland grant 13286580. Academy of Finland was not involved in the research beyond financial support. Otherwise this research did not receive any specific grant from funding agencies in the public, commercial, or not-for-profit sectors. The authors wish to thank Anders Glent Buch from the University of Southern Denmark for providing the CAD models of the Cranfield benchmark.

Sierla, S., Kyrki, V., Aarnio, P., Vyatkin, V. (2018) "Automatic assembly planning based on digital product descriptions", Computers in Industry, 97, pp. 34–46 https://doi.org/10.1016/j.compind.2018.01.013

## Glossary

Assembly Path Planning (APP): a sub-process of AP, in which collision-free paths for adding parts to a subassembly are computed

Assembly Planning (AP):the process of creating a detailed assembly plan to craft a whole product from separate parts considering the final product geometry, available resources, tool descriptions, etc.

Assembly Sequence Planning (ASP): a sub-process of AP, in which a sequence of collision-free operations is computed for bringing assembly parts together

AutomationML (Automation Markup Language) is a neutral data format based on XML for the storage and exchange of plant engineering information

CAEX (Computer Aided Engineering Exchange) is a neutral data format that allows storage of hierarchical object information, e.g. the hierarchical architecture of a plant. CAEX has been published as part of the IEC 62424. CAEX serves as top-level data format for the new neutral data exchange format AutomationML.

Digital twin: a near-real-time digital image of a physical object or process that helps optimize business performance

Product-centric control: an approach for materials handling and control, customization, and information sharing in the supply chain based on the unique identification of physical objects to which control instructions are then linked

XML (Extensible Markup Language) is a markup language that defines a set of rules for encoding documents in a format that is both human-readable and machine-readable